\documentclass{article}

\usepackage[numbers]{natbib}
\usepackage[final]{neurips_2019}

\usepackage[utf8]{inputenc} 
\usepackage[T1]{fontenc}   
\usepackage{url}           
\usepackage{booktabs}      
\usepackage{amsfonts}      
\usepackage{nicefrac}      
\usepackage{microtype}     
\usepackage[pagebackref=true,breaklinks=true,colorlinks,bookmarks=false]{hyperref}
\hypersetup{linkcolor=[rgb]{0.72,0.1,0.1}}
\hypersetup{citecolor=[rgb]{0.3,0.2,0.8}}
\hypersetup{urlcolor=[rgb]{0.36, 0.15, 0}}
\usepackage{graphicx} 
\usepackage{subfigure}
\expandafter\def\csname ver@subfig.sty\endcsname{}
\usepackage{wrapfig}
\usepackage{xspace}
\usepackage{caption}
\usepackage{svg}
\usepackage{algorithm}
\usepackage[noend]{algpseudocode}
\usepackage{tikz}
\usepackage{Definitions}
\usepackage{color}

\usetikzlibrary{bayesnet}

\renewcommand{\ge}{\geqslant}
\renewcommand{\citet}{\cite}
\renewcommand{\citep}{\cite}

\def\NoNumber#1{{\def\alglinenumber##1{}\State #1}\addtocounter{ALG@line}{-1}}

\title{Meta Architecture Search}

\author{
	Albert Shaw$^1$\thanks{Corresponding author: \texttt{ashaw596@gatech.edu}}\quad Wei Wei$^{2}$\quad Weiyang Liu$^1$\quad Le Song$^{1,3}$\quad Bo Dai$^{1, 2}$\\
	$^1$Georgia Institute of Technology \quad $^2$Google Research\quad $^3$Ant Financial\\
}

\begin{document}
	\maketitle
	\setcounter{footnote}{0}
	\begin{abstract}
		Neural Architecture Search~(NAS) has been quite successful in constructing state-of-the-art models on a variety of tasks. Unfortunately, the computational cost can make it difficult to scale. In this paper, we make the first attempt to study Meta Architecture Search which aims at learning a task-agnostic representation that can be used to speed up the process of architecture search on a large number of tasks. We propose the Bayesian Meta Architecture SEarch~(BASE) framework which takes advantage of a Bayesian formulation of the architecture search problem to learn over an entire set of tasks simultaneously. We show that on \texttt{Imagenet} classification, we can find a model that achieves 25.7\% top-1 error and 8.1\% top-5 error by adapting the architecture in less than an hour from an 8 GPU days pretrained meta-network. By learning a good prior for NAS, our method dramatically decreases the required computation cost while achieving comparable performance to current state-of-the-art methods - even finding competitive models for unseen datasets with very quick adaptation. We believe our framework will open up new possibilities for efficient and massively scalable architecture search research across multiple tasks\footnote{The code repository is available at \href{https://github.com/ashaw596/meta_architecture_search}{\url{https://github.com/ashaw596/meta\_architecture\_search}}.}.
	\end{abstract}
	
	\section{Introduction}
	
	For deep neural networks, the particular structure often plays a vital role in achieving state-of-the-art performance in many practical applications, and there has been much work~\cite{LeCBen15,HeZhaRenSun016,HuaSunLiuSedetal16,zhang1707shufflenet,liu2017spherenet,Liu2018DCNets,liu2019NSL,szegedy2015going,simonyan2014very,xie2019exploring} exploring the space of neural network designs.
    Due to the combinatorial nature of the design space, hand-designing architectures is time-consuming and inevitably sub-optimal. Automated Neural Architecture Search~(NAS) has had great success in finding high-performance architectures. However, people may need optimal architectures for several similar tasks at once, such as solving different classification tasks or even optimizing task networks for both high accuracy and efficient inference on multiple hardware platforms~\cite{FBNET}. Although there has been success in transferring architectures across tasks~\citep{transferable}, recent work has increasingly shown that the optimal architectures can vary between even similar tasks; to achieve the best results, NAS would need to be repeatedly run for each task~\citep{PROXYLESS} which can be quite costly.
	
\begin{figure}
		\includegraphics[width=0.87\linewidth]{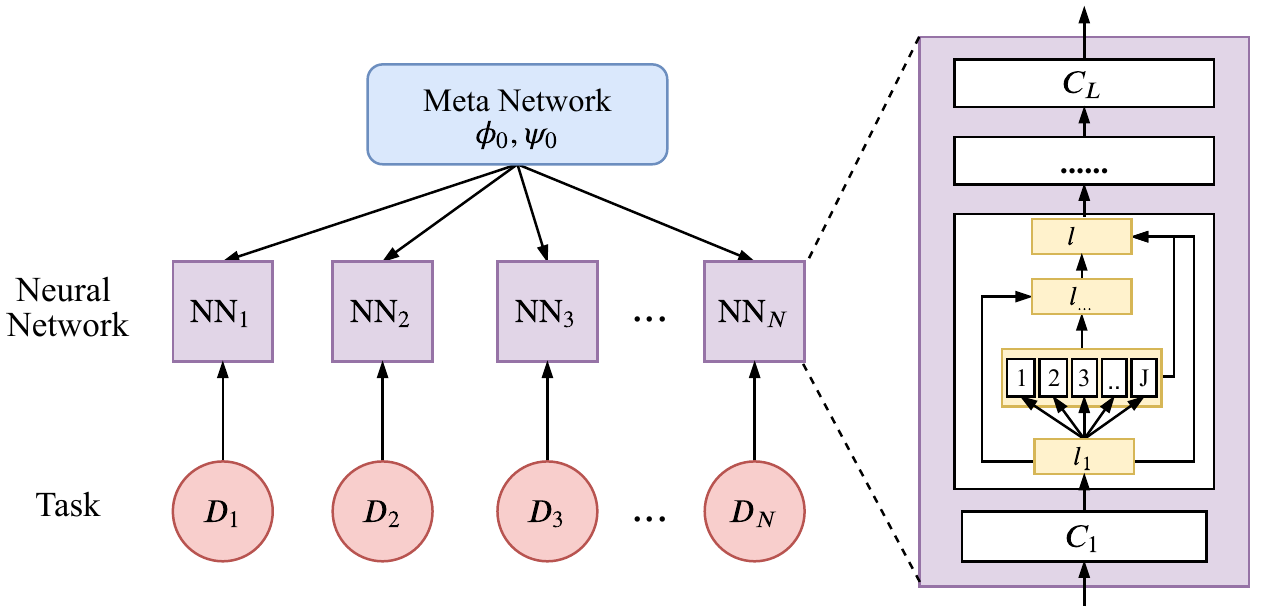}\\
		\vspace{-6mm}
		\centering
		\caption{Illustrations of Meta Architecture Search. We train a shared distribution for the meta-network and a sample from  the distribution will quick adapt to new task.}
		\label{fig:graphical_illustration}
		\vspace{-3mm}
	\end{figure}

	In this work, we present a first effort towards Meta Architecture Search, which aims at learning a task-agnostic representation that can be used to search over multiple tasks efficiently. The overall graphical illustration of the model can be found in Figure~\ref{fig:graphical_illustration}, where the meta-network represents the collective knowledge of architecture search across tasks. Meta Architecture Search takes advantage of the similarities among tasks and the corresponding similarities in their optimal networks, reducing the overall training time significantly and allowing fast adaptation to new tasks. We formulate the Meta Architecture Search problem from a Bayesian perspective and propose Bayesian Meta Architecture SEarch~(BASE), a novel framework to derive a variational inference method to learn optimal weights and architectures for a task distribution. To parameterize the architecture search space, we use a stochastic neural network which contains all the possible architectures within our architecture space as specific paths within the network. By using the Gumbel-softmax~\citep{jang2017categorical} distribution in the parameterization of the path distributions, this network containing an entire architecture space can be optimized differentially. To account for the task distribution in the posterior distribution of the neural network architecture and weights, we exploit the optimization embedding\citep{DaiDaiHeLiuetal18} technique to design the parameterization of the posterior. This allows us to train it as a meta-network optimized over a task distribution.
	
	To train our meta-network over a wide distribution of tasks with different image sizes, we define a new space of classification tasks by randomly selecting 10 \texttt{Imagenet}~\cite{DenDonSocLiEtal09} classes and downsampling the images to 32$\times$32, 64$\times$64, or 224$\times$224 image sizes. By training on these datasets, we can learn good distributions of architectures optimized for different image sizes. With a meta-network trained for 8 GPU days, we then show that we can achieve very competitive results on full \texttt{Imagenet} by deriving optimal task-specific architectures from the meta-network, obtaining 25.7\% top-1 error on \texttt{ImageNet} using an adaption time of less than one hour. Our method achieves significantly lower computational costs compared to current state-of-the-art NAS approaches. By adapting the multi-task meta-network for to the unseen \texttt{CIFAR10} dataset for less than one hour, we found a model that achieves 2.83\% Top-1 Error. Additionally, we also apply this method to tackle neural architecture search for few-shot learning, demonstrating the flexibility of our framework.

	Our research opens new potentials for using Meta Architecture Search across massive amounts of tasks. The nature of the Bayesian formulation makes it possible to learn over an entire collection of tasks simultaneously, bringing additional benefits such as computational efficiency and privacy when performing neural architecture search.
	
	\section{Related Work}
	
	\paragraph{Neural Architecture Search} 
	Several evolutionary and reinforcement learning based algorithms have been quite successful in achieving state-of-the-art performances on many tasks~\cite{zoph2016neural,transferable,real2018regularized,mobilenetv3}. However, these methods are computationally costly and require tremendous amounts of computing resources. While previous work has achieved good results with sharing architectures across tasks~\citep{transferable}, \citep{FBNET} and \cite{PROXYLESS} show that task and even platform-specific architecture search is required in order to achieve the best performance. Several methods~\cite{DARTS,ENAS,cai2018path,EAS,autodeeplab} have been proposed to reduce the search time, and both FBNet~\citep{FBNET} and SNAS~\citep{SNAS} utilize the Gumbel-Softmax~\citep{jang2017categorical} distribution similarly to our meta-network design to allow gradient-based architecture optimization. \citet{SMASH} and \citet{HYPER} also both propose methods to generate optimal weights for one task given any architecture like our meta-network is capable of. Their methods, however, do not allow optimization of the architectures and are only trained on a single task making them inefficient in optimizing over multiple tasks. Similarly to our work, \citep{DBLP:journals/corr/abs-1903-03536} recently proposed methods to accelerate search utilizing knowledge from previous searches and predicting posterior distributions of the optimal architecture. Our approach, however, achieves much better computational efficiency by not limiting ourselves to transferring knowledge from only the performance of discrete architectures on the validation datasets, but instead sharing knowledge for both optimal weights and architecture parameters and implicitly characterizing the entire dataset utilizing optimization embedding.
	
	\paragraph{Meta Learning}
	Meta-learning methods allow networks to be quickly trained on new data and new tasks~\cite{MAML,ravi2017optimization}.
    While previous works have not applied these methods to Neural Architecture Search, our derived Bayesian optimization method bears some similarities to Neural Processes~\citep{pmlr-v80-garnelo18a, DBLP:journals/corr/abs-1807-01622, kim2018attentive}. Both can derive a neural network specialized for a dataset by conditioning the model on some samples from the dataset. The use of neural networks allows both to be optimized by gradient descent. However, Neural Processes use specially structured encoder and aggregator networks to build a context embedding from the samples. We use the optimization embedding technique~\cite{DaiDaiHeLiuetal18} to condition our neural network using gradient descent in an inner loop, which allows us to avoid explicitly summarizing the datasets with a separate network. This inner-outer loop dynamic shares some similarities to second-order MAML~\citep{MAML}. Both algorithms unroll the stochastic gradient descent step. Due to this, we are also able to establish a connection between the heuristic MAML algorithm and Bayesian inference.

	\section{A Bayesian Inference View of Architecture Search}\label{sec:bayes_view}

	In this section, we propose a Bayesian inference view for neural architecture search which naturally introduces the hierarchical structures across different tasks. Such a view inspires an efficient algorithm which can provide a task-specific neural network with adapted weights and architecture using only \emph{a few} learning steps.
	
	We first formulate the neural architecture search as an operation selection problem. Specifically, we consider the neural network as a composition of $L$ layers of cells, where the cells share the same architecture, but have different parameters. In the $l$-th layer, the cell consists of a $K$-layer sub-network with bypass connections. Specifically, we denote the $x_k^l$ as the output of the $k$-th layer of $l$-th cell  
	\vspace{-0mm}
	\begin{equation}\label{eq:cell_selection}
	x_{k}^l = \sum_{i=1}^{k-1} \rbr{{z_{i, k}}^\top \Acal_{i}\rbr{\theta^l_{i,k}} }\circ x^l_i \defeq \sum_{i=1}^{k-1}\sum_{j=1}^J z_{ij,k} \phi_{i}^j\rbr{x^l_i; \theta^l_{ij,k}}
	\vspace{-0.5mm} 
	\end{equation}
	\vspace{-0.5mm} 
	where {
	\small
	$\Acal_i\rbr{\theta^l_{i,k}} = \sbr{\phi_i^j\rbr{\cdot; \theta^l_{ij,k}}}_{j=1}^J$} denotes a group of $J$ different operations from $\RR^d\rightarrow \RR^p$ which depend on parameters {\small $\theta^l_{ij,k}$}, \eg, different nonlinear neurons, convolution kernels with different sizes, or other architecture choices. $z_{i, k}$ are all binary variables which are shared across $L$ layers. They indicate which layers from the $1$ to $k-1$ levels in $l$-th cell should be selected as inputs to the $k$-th layer. Therefore, with different instantiations of $z$, the cell will select different operations to form the output. Figure~\ref{fig:graphical_illustration}
	has an illustration of this structure.
	
	We assume the probabilistic model as 
	\begin{equation}\label{eq:model}
	\begin{aligned}
	\theta^l_k := \sbr{\theta_{ij, k}^l}_{i, j=1}^{k-1, J}&\sim\Ncal\rbr{\mu^l_k, \rbr{\sigma^l_k}^2}, \\
	z_{i,k}&\sim \Ccal ategorial\rbr{\alpha_{i, k}},\,\, k = 1,\ldots, K,\\
	y&\sim p\rbr{y|x; \theta, z} \propto \exp\rbr{-\ell\rbr{f\rbr{x; \theta, z}, y}},
	\end{aligned}
	\end{equation}
	\vspace{0.5mm}
	with {\small $\theta = \cbr{[\theta^l_k]_{l=1}^L}_{k=1}^K$}, {\small $z = \cbr{[z_{i,k}]_{i=1}^{k-1}}_{k=1}^K$}, and {\small $\alpha_{i, k}^l \ge 0$}, {\small $\sum_{l=1}^L \alpha_{i, k}^l = 1$}. With this probabilistic model, the selection of $z$, \ie, neural network architecture search, is reduced to finding a distribution defined by $\alpha$, and the neural network learning is reduced to finding $\theta$, both of which are the parameters of the probabilistic model. 
	
	The most natural choice here for probabilistic model estimation is the maximum log-likelihood estimation~(MLE), \ie, 
	\begin{equation}\label{eq:mle}
	\textstyle
	\max_{W\defeq \rbr{\mu, \sigma, \alpha}}\,\, \widehat\EE_{x, y}\sbr{\log \int p\rbr{y|x; \theta, z}p\rbr{z;\alpha}p\rbr{\theta; \mu, \sigma}dzd\theta}. 
	\end{equation}
	However, the MLE is intractable due to the integral over latent variable $z$.
	We apply the classic variational Bayesian inference trick, which leads to the evidence lower bound~(ELBO), \ie, 
	\begin{equation}\label{eq:elbo}
	\textstyle
	\max_{W}\max_{q\rbr{z}, q\rbr{\theta}}\,\,-\widehat\EE_{x, y}{\EE_{z\sim q\rbr{z}, \theta\sim q\rbr{\theta}}}{\sbr{\ell\rbr{f\rbr{x; \theta, z}, y}} - KL\rbr{q(z)q(\theta)||p\rbr{z, \theta}}},
	\end{equation}
	where {\small $p\rbr{z} = \prod_{k=1}^K\prod_{i=1}^{k-1}\Ccal ategorial\rbr{z_{i,k}} = \prod_{k=1}^K \prod_{i=1}^{k-1}\prod_{l=1}^L \rbr{\alpha_{i, k}^l}^{z_{i, k}^l}$}. As shown in~\citet{Zellner88}, the optimal solution of~\eqref{eq:elbo} in all possible distributions will be the posterior. With such a model, architecture learning can be recast as Bayesian inference.

	\subsection{Bayesian Meta Architecture Learning}\label{subsec:bayes_meta}
	
	Based on the Bayesian view of architecture search, we can easily extend it to the meta-learning setting, where we have many tasks, \ie, $\Dcal_t =\cbr{x_i^t, y_i^t}_{i=1}^n$. {We are required to learn the neural network architectures and the corresponding parameters jointly while taking the task dependencies on the neural network structure into account. }
	
	We generalize the model~\eqref{eq:model} to handle multiple tasks as follows. For the $t$-th task, we design the model following~\eqref{eq:model}. Meanwhile,
	the hyperparameters, \ie, $\rbr{\mu, \sigma, \alpha}$, are shared across all the tasks. In other words, the layers and architecture priors are shared between tasks. Then we have the MLE:
	\vspace{-1mm}
	\begin{equation}\label{eq:meta_mle}
	\max_{W}\,\, \widehat\EE_{\Dcal_t}\widehat\EE_{\rbr{x, y}\sim \Dcal_t}\sbr{\log \int p\rbr{y|x; \theta, z}p\rbr{z;\alpha}p\rbr{\theta; \mu, \sigma}dzd\theta}
	\end{equation}
	Similarly, we exploit the ELBO. Due to the structures induced by sharing across the tasks, the posteriors for $\rbr{z, \theta}$ have special dependencies, \ie, 
	
	\begin{equation}\label{eq:meta_elbo}
	\max_{W}\widehat\EE_{\Dcal_t}\rbr{\max_{q\rbr{z|\Dcal}, q\rbr{\theta|\Dcal}}\,\,{\widehat\EE_{\rbr{x, y}\sim\Dcal_t}\EE_{z\sim q\rbr{z|\Dcal},\theta\sim q\rbr{\theta|\Dcal}}\sbr{-\ell\rbr{f\rbr{x; \theta, z}, y}} - KL\rbr{q||p}} }
	\end{equation}
	
	With the variational posterior distributions, $q\rbr{z|\Dcal}$ and $q\rbr{\theta|\Dcal}$, introduced into the model, we can directly generate the architecture and its corresponding weights based on the posterior. In a sense, the posterior can be understood as the neural network predictive model.

	\section{Variational Inference by Optimization Embedding}\label{sec:vi_opt}

	The design of the parameterization of the posterior {\small $q\rbr{z|\Dcal}$} and {\small$q\rbr{\theta|\Dcal}$} is extremely important, especially in our case where we need to model the dependence between {\small$\rbr{z, \theta}$} w.r.t. the \emph{task distributions} {\small$\Dcal$} and the \emph{loss information}. Fortunately, we can bypass this problem by applying parameterized Coupled Variational Bayes~(CVB), which generates the parameterization automatically through \emph{optimization embedding}~\citep{DaiDaiHeLiuetal18}. 
	
	Specifically, we assume the {\small$q\rbr{\theta|\Dcal}$} is Gaussian and the {\small$q\rbr{z|\Dcal}$} is a product of the categorical distribution. We approximate the categorical $z$ with the Gumbel-Softmax distribution~\cite{jang2017categorical,MadMniTeh16}, which leads to a valid gradient so that the model will be fully differentiable. Therefore, we have
	\vspace{-2mm}
	\begin{equation}\label{eq:posterior}
		q_\psi\rbr{\theta|\Dcal} =\Ncal\rbr{\psi_\mu, \psi_\sigma},\quad
		q_\phi\rbr{z_{i, k}|\Dcal} = \Gamma\rbr{r}\tau^{L-1}\rbr{\sum_{l=1}^L \frac{\pi_{\Dcal, \phi_{i,k}^l, l}}{\rbr{z_{i, k}^l}^\tau}}^{-r}\prod_{i=1}^r \rbr{\frac{\pi_{\Dcal, \phi_{i, k}^l, l}}{\rbr{z_{i, k}^l}^{\tau+1}}}
	\vspace{-2mm}
	\end{equation}
	Then, we can sample {\small$\rbr{\theta, z}$} by following,\\
	\begin{equation}\label{eq:posterior_param}
		\begin{aligned}
		\theta_\Dcal\rbr{\epsilon, \psi} &= \psi_{\Dcal, \mu} + \epsilon \psi_{\Dcal, \sigma}, \quad \epsilon\sim\Ncal\rbr{0, 1}, \\
		z_{i, k,\Dcal}^l\rbr{\xi, \phi} &= \frac{\exp\rbr{\rbr{\phi_{\Dcal, i, k}^l + \xi^l} / \tau}}{\sum_{l=1}^L \exp\rbr{\rbr{\phi_{i, k}^l + \xi^l} / \tau}}, \quad \xi^l \sim \Gcal\rbr{0, 1},\quad l\in\cbr{1, \ldots, L},
		\end{aligned}
	\end{equation}
	with {\small$\pi_{x, \phi, i} = \frac{\exp\rbr{\phi_{x, i}}}{\sum_{i=1}^p \exp\rbr{\phi_{x, i}}}$} and {\small$\Gcal\rbr{0, 1}$} denotes the Gumbel distribution. We emphasize that we do not have any explicit form of the parameters {\small$\phi_\Dcal$} and {\small$\psi_\Dcal$} yet, which will be derived by optimization embedding automatically.

	Plugging the formulation into the ELBO~\eqref{eq:meta_elbo}, we arrive at the objective

	\begin{equation}
	\label{eq:param_obj}
		\widehat\EE_{\Dcal}\Big[{\max_{\phi_\Dcal, \psi_\Dcal} \underbrace{\widehat\EE_{x, y}\EE_{\xi, \epsilon}\sbr{-\ell\rbr{f\rbr{x; \theta_\Dcal\rbr{\epsilon, \psi}, z_\Dcal\rbr{\xi, \phi}}, y}} -\log \frac{q_\phi\rbr{z|\Dcal}}{p\rbr{z;\alpha}} - \log\frac{q_\psi\rbr{\theta|\Dcal}}{p\rbr{\theta; \mu, \sigma}} }_{L\rbr{\phi_\Dcal, \psi_\Dcal; W}}}\Big].
	\end{equation}
	With the ultimate objective~\eqref{eq:param_obj} we follow the parameterized CVB derivation~\citep{DaiDaiHeLiuetal18} for embedding the optimization procedure for {\small $\rbr{\phi,\psi}$}. Denoting the {\small $\ghat_{\phi_\Dcal, \psi_\Dcal}\rbr{\Dcal, W} = \frac{\partial \Lhat}{\partial \rbr{\phi_\Dcal, \psi_\Dcal}}$} where {\small$\Lhat$} is the stochastic approximation for {\small$L\rbr{\phi_\Dcal, \psi_\Dcal; W}$}, then, the stochastic gradient descent~(SGD) iteratively updates as 
	\begin{equation}\label{eq:unrolling}
	\textstyle
	\sbr{\phi^t_\Dcal, \psi^t_\Dcal} = \eta_t \ghat_{\phi_\Dcal, \psi_\Dcal}\rbr{\Dcal, W} + \sbr{\phi^{t-1}_\Dcal, \psi^{t-1}_\Dcal},
	\end{equation}
	We can initialize {\small $\rbr{\phi^0, \psi^0} = W$} which is shared across all the tasks. Alternative choices are also possible, \eg, with one more neural network, {\small$\rbr{\phi^0, \psi^0} = h_V\rbr{\Dcal}$}. We unfold $T$ steps of the iteration to form a neural network with output {\small$\rbr{\phi^T_\Dcal, \psi^T_\Dcal}$}. Plugging the obtained {\small$\rbr{\phi^T_\Dcal, \psi^T_\Dcal}$} to~\eqref{eq:posterior_param}, we have the parameters and architecture as {\small$\rbr{\theta^T_\Dcal\rbr{\xi, \psi^T_\Dcal}, z_\Dcal\rbr{\xi, \phi^T_\Dcal}}$}. In other words, we derive the concrete parameterization of {\small$q\rbr{\theta|\Dcal}$} and {\small$q\rbr{z|\Dcal}$} automatically by unfolding the optimization steps. Replacing the parameterization of {\small$q\rbr{z|\Dcal}$} and {\small$q\rbr{\theta|\Dcal}$} into {\small$L\rbr{\phi_\Dcal, \psi_D, W}$}, we have 
	\begin{equation}\label{eq:ultimate_obj}
    	\max_{W}\,\, \widehat\EE_{\Dcal}{ {\widehat\EE_{x, y}\EE_{\xi, \epsilon}\Big[{\underbrace{-\ell\rbr{f\rbr{x; \theta^T_\Dcal\rbr{\epsilon, \psi}, z^T_\Dcal\rbr{\xi, \phi}}, y} -\log \frac{q_{\phi^T_\Dcal}\rbr{z|\Dcal}}{p\rbr{z;\alpha}} - \log\frac{q_{\psi^T_\Dcal}\rbr{\theta|\Dcal}}{p\rbr{\theta; \mu, \sigma}}}_{\Lhat\rbr{x, y, \epsilon, \xi; W}} }}}\Big].
    	\vspace{-3mm}
	\end{equation}
	
	\begin{wrapfigure}{r}{0.65\textwidth}
		\vspace{-6mm}
		\begin{minipage}{0.65\textwidth}
			\begin{algorithm}[H]
				\caption{Bayesian meta Architecture SEarch~(BASE)}\label{fig:psuedo_code}
				\begin{algorithmic}[1] 
					\State Initialize meta-network parameters $W_0$.
					\For{$e = 1,\ldots, E$}
						\State Sample $C$ tasks $\cbr{\Dcal_c}_{c=1}^C \sim \Dcal$.
						\For{$\Dcal_c$ in $\Dcal$}
						\State Sample $\cbr{x_t, y_t}_{t=1}^T\sim\Dcal_c$.
						\State Let $\phi^0_{c}, \psi^0_{c} = W_{e-1}$.
						\For{$t = 1,\ldots, T$}
						\State Sample $\xi\sim\Gcal\rbr{0, 1}$.
						
						\State Update $\sbr{\phi^t_{c}, \psi^t_{c}} =\sbr{\phi^{t-1}_{c}, \psi^{t-1}_{c}} - $
						\NoNumber{\hspace{\algorithmicindent}$\eta\nabla_{\phi^{t-1}_{c}, \psi^{t-1}_{c}} \Lhat(f(x_t; \phi^{t-1}_{c}, \psi^{t-1}_{c}, \xi), y_t) $.}
						\EndFor
					\EndFor
					\State Update $W_{e} = W_{e-1} + \lambda\frac{1}{C}\sum_{c=1}^C (\sbr{\phi^T_{c}, \psi^T_{c}} - W_{e-1})$.
					\EndFor
				\end{algorithmic}
			\end{algorithm}
		\end{minipage}
	\end{wrapfigure}
	If we apply stochastic gradient ascent in the optimization~\eqref{eq:ultimate_obj} for updating $W$, the instantiated algorithm from optimization embedding shares some similarities to second-order MAML~\citep{MAML} and DARTS~\citep{DARTS} algorithms. Both of these two algorithms unroll the stochastic gradient step. However, with the introduction of the Bayesian view, we can exploit the rich literature for the approximation of the distributions on discrete variables. More importantly, we can easily share both the architecture and weights across many tasks. Finally, this establishes the connection between the heuristic MAML algorithm to Bayesian inference, which can be of independent interest. 
    
    \noindent{\bf Practical algorithm:} In the method derivation, for the simplicity of exposition, we assumed there is only one cell shared across all the layers in every task, which may be overly restrictive. Following \citet{transferable}, we design two types of cells, named as a normal cell with $\phi_{n}$ and a reduction cell with $\phi_r$, which appear alternatively in the neural network. Please refer to Appendix~\ref{appendix:cell_parameters} for an illustration.
    
    In practice, the multistep-unrolling of the gradient computation is expensive and memory inefficient. We can exploit the finite difference approximation for the gradient. This is similar to the iMAML~\citep{rajeswaran2019metalearning} and REPTILE~\citep{REPTILE} approximations of MAML. Moreover, we can further accelerate learning by exploiting parallel computation. Specifically, for each task, we start from a local copy of the current $W$ and apply stochastic gradient ascent based on the task-specific samples. Then, the shared $W$ can be updated by summarizing the task-specific parameters and architecture. The pseudo-code for the concrete algorithm for Bayesian meta-Architecture SEarch~(BASE) can be found in Algorithm~\ref{fig:psuedo_code}.
	
	With a meta-network trained with BASE over a series of tasks, for a new task, we can adapt an architecture by sampling from the posterior distribution of $z_D$ through~\eqref{eq:posterior} with $\sbr{\phi^T_D, \psi^T_D}$ calculated by~\eqref{eq:unrolling} given new task $D$ which will be used to define the full-sized network.
    Illustrations of the network motifs used for the search network and the full networks can be found in Appendix~\ref{appendix:single_task_arch}. More details about the architecture space can be found in Appendix~\ref{appendix:arch_space}.

	\section{Experiments and Results}
	
	\subsection{Experiment Setups}
	
	\paragraph{Downsampled Multi-task Datasets}
	To help the meta-network generalize to inputs with different sizes, we create three new multi-task datasets: \texttt{Imagenet32}(\texttt{Imagenet} downsampled to 32x32), \texttt{Imagenet64}(\texttt{Imagenet} downsampled to 64x64), and \texttt{Imagenet224}(\texttt{Imagenet} downsampled to 224x224). \texttt{Imagenet224} uses the most commonly used size for inference for the full \texttt{Imagenet} dataset in the mobile setting. Our tasks are defined by sampling 10 random classes from one of the resized \texttt{Imagenet} datasets similar to the \texttt{Mini-Imagenet} dataset~\cite{vinyals2016matching} in few-shot learning. This allows us to sample tasks from a space of $\thickmuskip=2mu \medmuskip=2mu C(1000,10) \times 3 \approx 2.634 \times 10^{23}$ tasks.
	
	\vspace{-2mm}
	\paragraph{Featurization Layers}
	To conduct architecture search on these multi-sized, multi-task datasets, the meta-network uses separate initial featurization layers (heads) for each image size.
    The use of non-shared weights for the initial image featurization both allows the meta-network to learn a better prior as well as enabling the use of different striding in the heads to compensate for the significant difference in image sizes.
    The \texttt{Imagenet224} head strides the output to 1/8th of the original input while the 32x32 and 64x64 heads both stride to 1/2th the original input size. 

	\subsection{Search Performance}
    We validated our meta-network by transferring the results of architectures optimized for \texttt{CIFAR10}, \texttt{SVHN}, and \texttt{Imagenet224} to full-sized networks. Details of how we trained the full networks can be found in Appendix~\ref{appendix:retraining_details}. To derive the full-sized \texttt{Imagenet} architectures, we select a high probability architectures from the posterior distribution of architectures given random 10-class \texttt{Imagenet224} datasets by averaging the sampled architecture distributions for 8 random datasets. To derive the \texttt{CIFAR10} and \texttt{SVHN} architectures, we adapted the network on the unseen datasets and selected the architecture with the highest probability of being chosen.
    The meta-network was trained for 130 epochs. At each epoch, we sampled and trained on a total of 24 tasks, sampling 8 10-class discrimination tasks each from \texttt{Imagenet32}, \texttt{Imagenet64}, and \texttt{Imagenet224}. 
    All experiments were conducted with Nvidia 1080 Ti GPUs.

	\begin{table*}
		\centering
		\caption{Classification Accuracies on \texttt{CIFAR10}}
		\begin{tabular}{llll}
			\toprule        
			Architecture & Top-1 Test & Parameters & Search Time \\
			&Error&(M)&(GPU Days)\\
			\midrule
			NASNet-A + cutout~\citep{transferable} & 2.65 & 3.3 & 1800 \\ 
			AmoebaNet-A + cutout \citep{real2018regularized}& $3.34 \pm 0.06$ & 3.2 & 3150 \\ 
			AmoebaNet-B + cutout~\citep{real2018regularized} & $\mathbf{2.55 \pm 0.05}$ & \textbf{2.8} & 3150 \\ 
			Hierarchical Evo~\citep{LiuSimVinFeretal17} & $3.75 \pm 0.12$ & 15.7 & 300 \\ 
			PNAS~\citep{PNAS} & $3.41 \pm 0.09$ & 3.2 & 225 \\ 
			\midrule
			DARTS (1st order bi-level) + cutout~\citep{DARTS} & $3.00 \pm 0.14$ & 3.3 & 1.5 \\
			DARTS (2nd order bi-level) + cutout~\citep{DARTS} & $\mathbf{2.76 \pm 0.09}$ & 3.3 & 4 \\
			SNAS (single-level)  + cutout~\citep{SNAS} & $2.85 \pm 0.02$ & \textbf{2.8} & 1.5 \\ 
			SMASH~\citep{SMASH} & 4.03 & 16 & 1.5\\
			ENAS + cutout~\citep{ENAS} & 2.89 & 4.6 & 0.5 \\ 
			
			\midrule
			BASE (Multi-task Prior) & $3.18$ & 3.2 & 8 Meta \\
			BASE (\texttt{Imagenet32} Tuned) & $3.00$ & 3.3 & 0.04 Adap / 8 Meta\\
			BASE (\texttt{CIFAR10} Tuned) & \textbf{2.83} & \textbf{3.1} & 0.05 Adap / 8 Meta\\
			\bottomrule
		\end{tabular}
		\label{cifar10-results-table}
	\end{table*}
	
    \paragraph{Performance on CIFAR10 Dataset}
    The result of our Meta Architecture Search on \texttt{CIFAR10} can be found in Table~\ref{cifar10-results-table}. We compared a few variants of our methods. BASE (Multi-task Prior) is architecture derived from training on the multi-task \texttt{Imagenet} datasets only without further fine-tuning. This model did not have access to any information on the \texttt{CIFAR10} dataset and is used as a baseline comparison. 
	
	The BASE (\texttt{Imagenet32} Tuned) is the network derived from the multi-task prior fine-tuned on \texttt{Imagenet32}. We chose \texttt{Imagenet32} since it has the same image dimension as \texttt{CIFAR10}. It does slightly better than the BASE (Multi-task Prior) on \texttt{CIFAR10}. We compare these networks to the BASE (\texttt{CIFAR10} Tuned), which is the network derived from the meta-network prior fine-tuned on \texttt{CIFAR10}. Not surprisingly, this network performs the best as it has access to both the multi-task prior and the target dataset. One thing to note is that for BASE (\texttt{Imagenet32} Tuned) and BASE (\texttt{CIFAR10} Tuned), we only fine-tuned the meta-networks for 0.04 GPU days and 0.05 GPU days respectively. The adaptation time required is significantly less than that required for the initial training of the multi-task prior, as well as the required search time for the rest of the baseline NAS algorithms. With respect to the number of parameters, our models are comparable in size with to the baseline models. Using adaptation from our meta-network prior, we can find high performing models while using significantly less compute.   
	
	\paragraph{Performance on \texttt{SVHN} Dataset}
	The result of our Meta Architecture Search on \texttt{SVHN} are shown in Table~\ref{svhn-results-table}. We used the same multi-task prior previously trained on the multi-scale \texttt{Imagenet} datasets and quickly adapted the meta-network to \texttt{SVHN} in less than an hour. We also trained the \texttt{CIFAR10} specialized architecture found in DARTS~\citet{DARTS}. The adapted network architecture achieves the best performance in our experiments and has comparable performance to other work for the model size. This also validates the importance of task-specific specialization since it significantly improved the network performance over both our multi-task prior and \texttt{Imagenet32} tuned baselines.

	\begin{table*}
		\centering
		\caption{Classification Accuracies on \texttt{SVHN}}
		\begin{tabular}{llll}
			\toprule        
			Architecture & Top-1 Test & Parameters & Search Time \\			&Error&(M)&(GPU Days)\\
			\midrule
			WideResnet~\cite{zagoruyko2016wide} & \textbf{1.30 $\pm$ 0.03} & 11.7\hphantom{0} & - \\
			MetaQNN~\cite{baker2016designing} & 2.24 & \textbf{9.8} & 100 \\
			\midrule
			DARTS~(\texttt{CIFAR10} Searched) & \textbf{2.09} & \textbf{3.3} & 4\\ 
			\midrule
			BASE~(Multi-task Prior) & 2.13 & 3.2 & 8 Meta\\
			BASE~(\texttt{Imagenet32} Tuned) & 2.07 & 3.3 & 0.04 Adap / 8 Meta\\
			BASE~(\texttt{SVHN} Tuned) & \textbf{2.01} &  \textbf{3.2} & 0.04 Adap / 8 Meta\\
			\bottomrule
		\end{tabular}
		\label{svhn-results-table}
	\end{table*}
	
	\vspace{-1.5mm}
	
	\begin{table*}
		\centering
	    \caption{Classification Accuracies on \texttt{Imagenet}}
		\begin{tabular}{llllll}
			\toprule        
			Architecture & Top-1 & Top-5 & Params & MACs & Search Time \\
			&Err&Err&(M)&(M)&(GPU Days)\\
			\midrule
			
			NASNet-A~\citep{transferable} & 26.0 & 8.4 & 5.3 & 564 & 1800
			\\NASNet-B~\citep{transferable} & 27.2 & 8.7 & 5.3 & \textbf{488} & 1800
			\\NASNet-C~\citep{transferable} & 27.5 & 9.0 & \textbf{4.9}~~ & 558 & 1800
			\\AmoebaNet-A~\citep{real2018regularized} & 25.5 & 8.0 & 5.1 & 555 & 3150
			\\AmoebaNet-B~\citep{real2018regularized} & 26.0 & 8.5 & 5.3 & 555 & 3150
			\\AmoebaNet-C~\citep{real2018regularized} & \textbf{24.3} & \textbf{7.6} & 6.4 & 570 & 3150
			\\PNAS~\citep{PNAS} & 25.8 & 8.1 & 5.1 & 588 & 225\\
			\midrule
			DARTS~\citep{DARTS} & \textbf{26.9} & \textbf{9.0} & 4.9 & 595 & 4 \\			
			SNAS~\citep{SNAS} & 27.3 & 9.2 & \textbf{4.3} & \textbf{522} & 1.5 \\ 
			\midrule
			BASE~(Multi-task Prior) & $26.1$ & $8.5$ & \textbf{4.6} & \textbf{544} & 8 Meta  \\
			BASE~(\texttt{Imagenet} Tuned) & \textbf{25.7} & \textbf{8.1} & 4.9 & 559 & 0.04 Adap / 8 Meta \\
			\bottomrule
		\end{tabular}
		  \vspace{-2.5mm}
		\label{imagenet-results-table}
	\end{table*}
	
	\begin{wrapfigure}{r}{0.48\textwidth}
		\centering
 		\vspace{-4mm}
\includegraphics[trim={4mm 0mm 4mm 20mm},clip=true, width=0.52\textwidth]{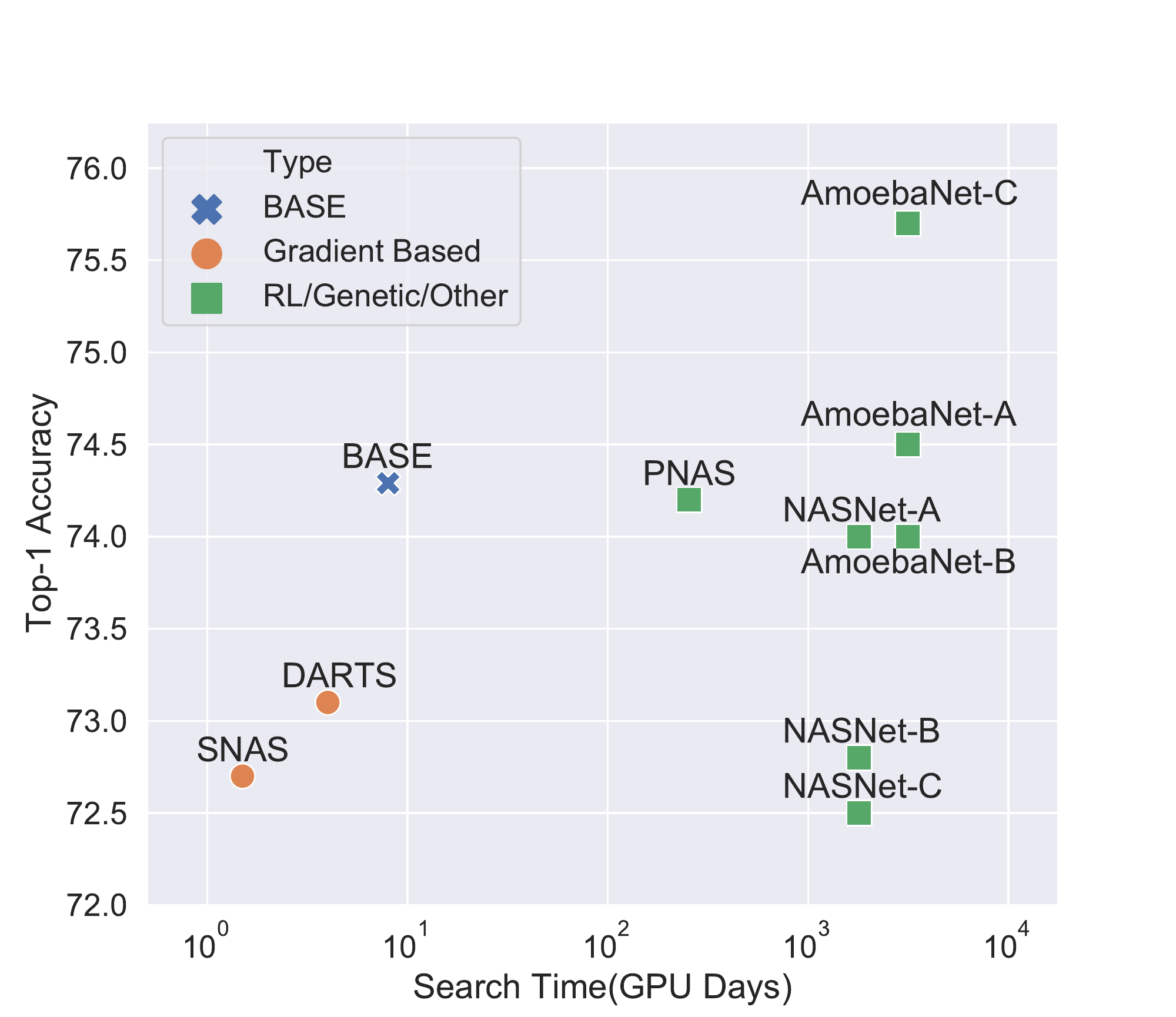}
		\vspace{-5.5mm}
		\caption{Top-1 \texttt{Imagenet} Accuracy vs Search Time in GPU Days of different NAS methods on \texttt{Imagenet}.}
		\label{fig:run_time_comparison}
		\vspace{1mm}
	\end{wrapfigure}
	
		\paragraph{Performance on \texttt{ImageNet} Dataset}
The results of our Meta Architecture Search on \texttt{Imagenet} can be found in Table~\ref{imagenet-results-table}. We compare BASE (Multi-task Prior) with Base (\texttt{Imagenet} Tuned), which is the multi-task prior tuned on 224x224 \texttt{Imagenet}. The performance of our Imagenet Tuned model actually exceeds that of existing differential NAS approaches DARTS~\cite{DARTS} and SNAS~\cite{SNAS} on both top-1 Error and top-5 error.  In terms of number of parameters and Multiply Accumulates(MAC), our found models are comparable to state-of-the-art networks. Considering running time, while the multi-task pretraining took 8 GPU days, we only needed 0.04 GPU days to adapt to full sized \texttt{Imagenet}. In Figure~\ref{fig:run_time_comparison}, we compare our models with other NAS approaches with respect to top-1 error and search time. For fairness, we include the time required to learn the architecture prior, and we still achieve significant accuracy gains for our computational cost. 

	\begin{figure}
		\hspace{-2mm}
		\centering
		\begin{tabular}{cc}
			\includegraphics[width=0.42\textwidth, trim={15mm 12mm 15mm 38mm},clip]{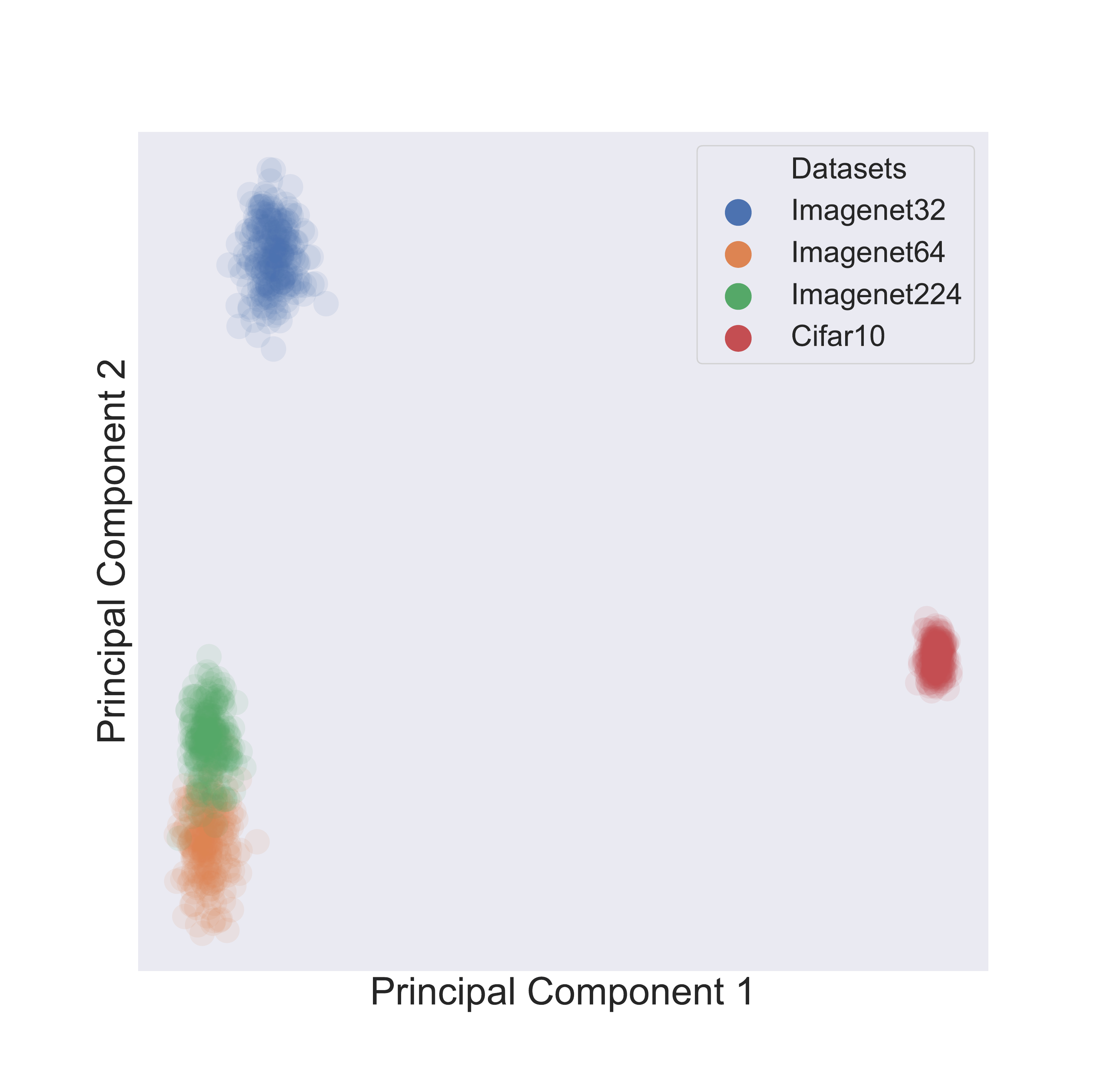}\vspace{-1mm}&
			\includegraphics[width=0.42\textwidth, trim={15mm 12mm 15mm 38mm},clip]{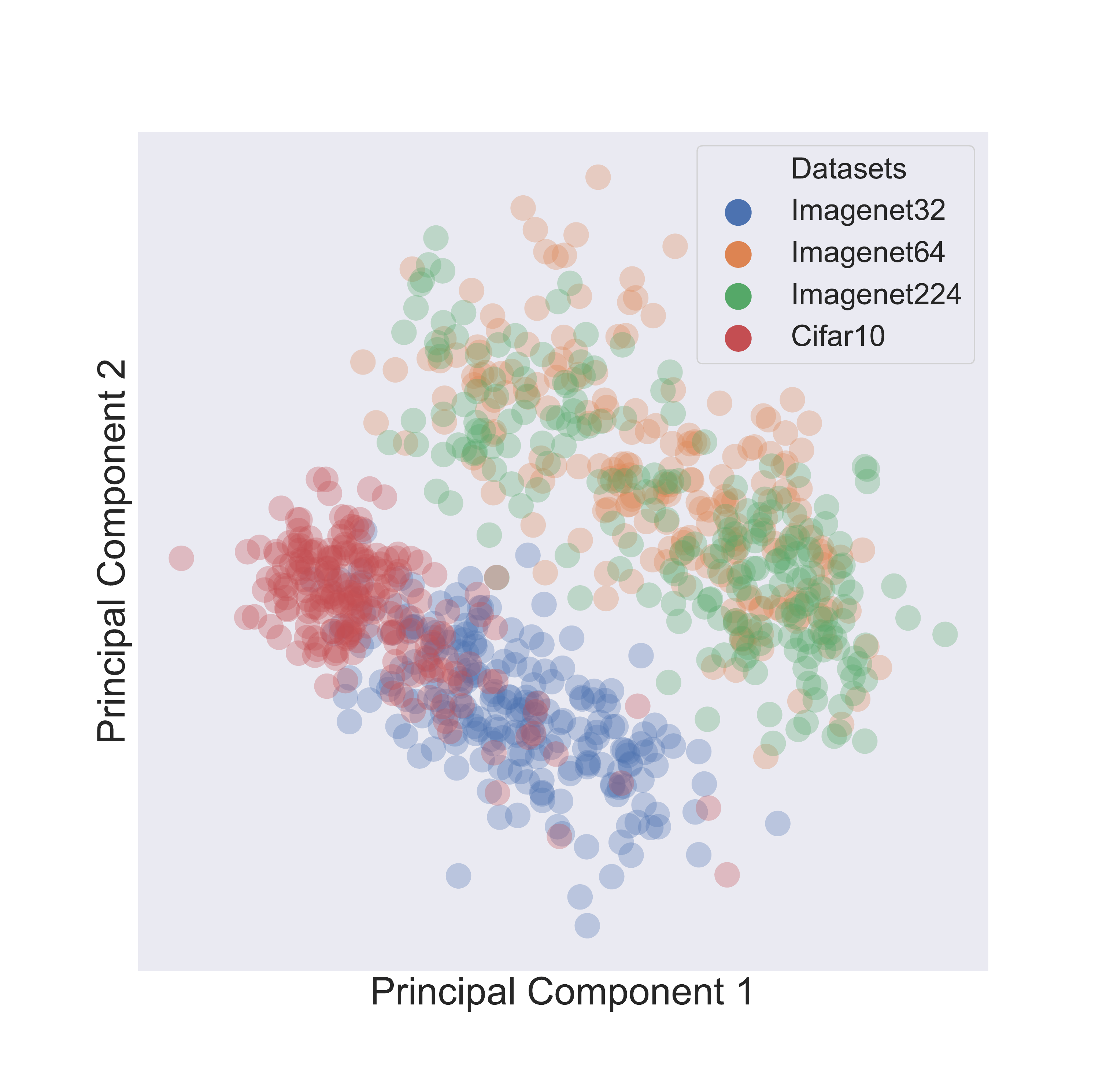}\vspace{-1mm}\\
		    (a) PCA of weights & (b) PCA of architecture\\
		\end{tabular}
		\caption{Visualization of the PCA for $\rbr{\theta,  z}$, \ie, weight and architecture, sampled from the posterior distribution of the meta-network.}\label{figure:architecture_pca}
		\vspace{-1mm}
    \end{figure}

	\section{Empirical Analysis}
In this section, we analyze the task-dependent parameter distributions derived from meta-network adaptation and demonstrate the abilities of the proposed method for fast adaptation as well as architecture search for few-shot learning.

	\subsection{Visualization of Posterior Distributions}
    Figure~\ref{figure:architecture_pca} shows the PCA visualization of the posterior distributions of the convolutional weights $\psi_D^t$ and architecture parameters $\phi_D^t$. The \texttt{CIFAR10} optimized distributions were derived by quick adapting the pretrained meta-network for the \texttt{CIFAR10} dataset while the other distributions were adapted for tasks sampled from the corresponding multi-task datasets. We see that the distribution of weights is more concentrated for \texttt{CIFAR10} than for other datasets, likely since it corresponds to a single task instead of a task distribution. It also seems that the \texttt{Imagenet224} and \texttt{Imagenet64} posterior weight and architecture distributions are close to each other. This is likely due to the fact they are the closest to each other in feature resolution after being strided down by the feature heads to $28\times 28$ and $32\times 32$. 
    Considering the visualization of the architecture parameter distributions, it's notable that while the closeness of clusters seems to indicate a similarity between \texttt{Imagenet32} and \texttt{CIFAR10}, \texttt{CIFAR10} still has a clearly distinct cluster. This seems to support that even though the meta-network prior was never trained on \texttt{CIFAR10}, an optimized architecture posterior distribution can be quickly derived for \texttt{CIFAR10}.
	
	\begin{wrapfigure}{r}{0.51\textwidth}
		\vspace{-5mm}
		\includegraphics[trim=0 0 0 16mm ,clip,width=0.53\textwidth]{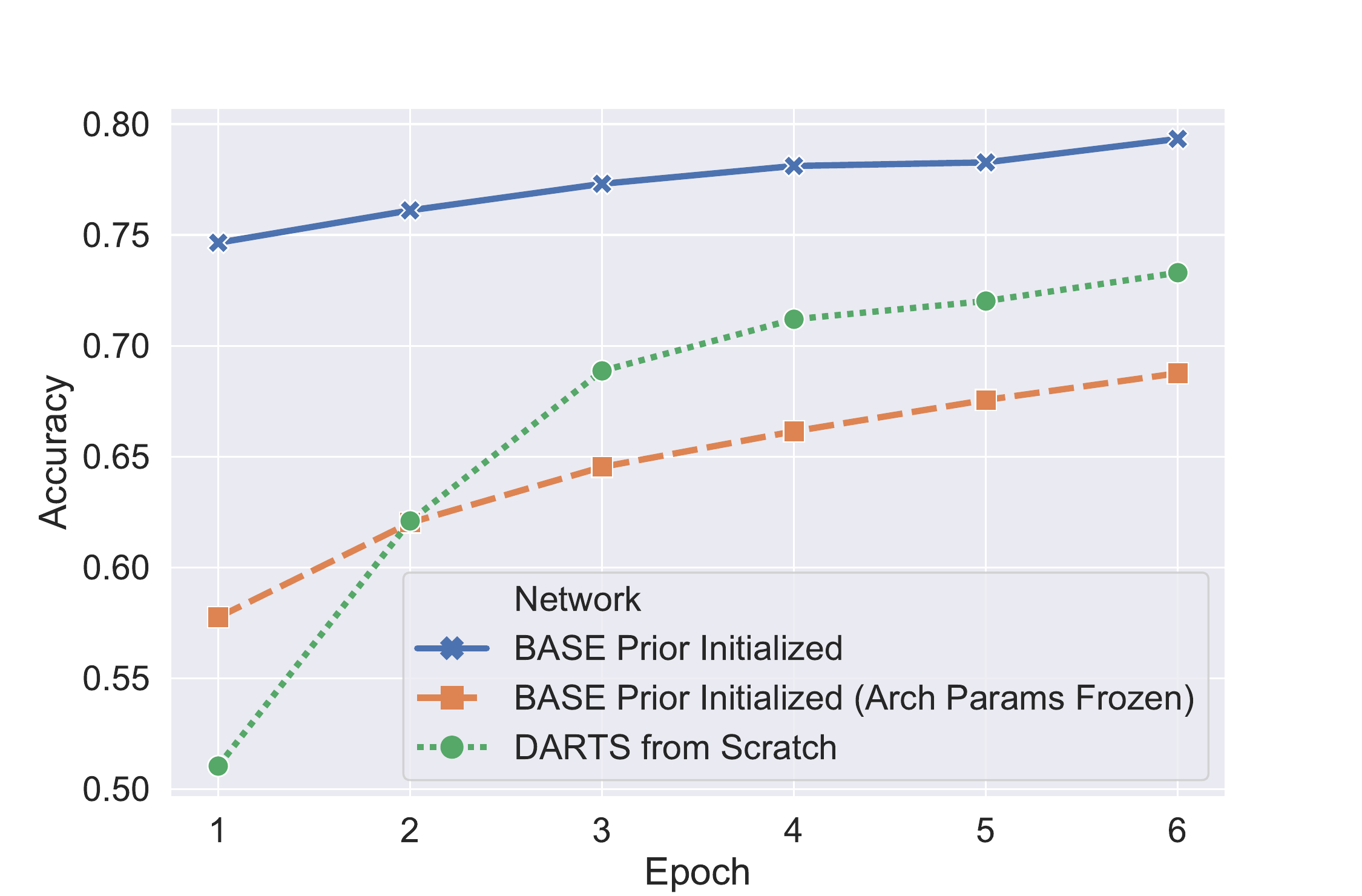}
		\vspace{-5mm}
		\caption{Graph showing the fast adaptation properties of pretrained meta-networks when adapting to \texttt{CIFAR10} in a few epochs.}
		\label{fig:fast_adaption}
		\vspace{-3mm}
	\end{wrapfigure}

	\subsection{Fast Adaptations}
	In this section, we explore the direct transfer of both architecture and convolutional weights from the meta-network by comparing the test accuracy we get on \texttt{CIFAR10} with meta-networks adapted for six epochs. The results are shown in Figure~\ref{fig:fast_adaption}. We compare against the baseline accuracy of the DARTS~\cite{DARTS} super-network trained from scratch on \texttt{CIFAR10}. Our meta-network adapted normally from a multi-task prior, achieves an accuracy of around $0.75$ after only one epoch. We also experimented with freezing the architecture parameters, which greatly degraded the performance. This shows the importance of co-optimizing both the weight and architecture parameters.
	\subsection{Few-Shot Learning}
	In order to show the generalizability of our algorithm, we used it to conduct an architecture search over the few-shot learning problem. Since few-shot learning targets adapting in very few samples, we can avoid using the Finite Difference approximation and directly use the optimization-embedding technique in these experiments. These experiments were run on a commonly used benchmark for few-shot learning, the \texttt{Mini-Imagenet} dataset as proposed in~\citet{vinyals2016matching}, specifically on the 5-way classification 5-shot learning problem.

	\begin{table*}
        \centering
        \caption{Comparison of few-shot learning baselines against MAML~\citep{MAML} using the architectures found by our BASE algorithm on few-shot learning on the \texttt{Mini-Imagenet}  dataset.}
		\label{few_shot_results}
			\begin{tabular}{llll}
				\toprule        
				Architecture & 5-shot Test & Params& Few-shot  \\
			    &Accuracy&(M)& Algorithm\\[0.5ex]
				\midrule
				MAML~\citep{MAML} & 63.11 $\pm$ 0.92\% & 0.1 & MAML\\
				REPTILE~\citep{REPTILE} & 65.99 $\pm$ 0.58\% & 0.1 & REPTILE\\
				\midrule
				DARTS Architecture & $63.95 \pm 1.1\%$ & 1.6 & MAML \\
				BASE~(Softmax) & $65.4 \pm 0.74\%$ & 1.2 & MAML \\ 
				\textbf{BASE~(Gumbel)} & \textbf{66.2 $\pm$ 0.7\%} & \textbf{1.2} & MAML\\
				\bottomrule
			\end{tabular}
			\vspace{2mm}
	\end{table*}
    The full-sized network is trained on the few-shot learning problem using second-order MAML~\citep{MAML}.
    Search and full training were run twice for each method. A variation of our algorithm was also run using a simple softmax approximation of the Categorical distribution as proposed in~\citet{DARTS} to test the effect of the Gumbel-Softmax architecture parameterization.
	The full results are shown in Table~\ref{few_shot_results}, our searched architectures achieved significantly better average testing accuracies than our baselines on five-shot learning on the \texttt{Mini-Imagenet}  dataset in the same architecture space. The \texttt{CIFAR10} optimized DARTS architecture also achieved results that were significantly better than that found in the original MAML baseline~\citep{MAML} showing some transferability between \texttt{CIFAR10} and meta-learning on \texttt{Mini-Imagenet}. That architecture, however, also had considerably more parameters than our found architectures and trained significantly slower. The Gumbel-Softmax meta-network parameterization also found better architectures than the simple softmax parameterization.
	
	\vspace{2mm}
	\section{Conclusion}

	In this work, we present a Bayesian Meta-Architecture search~(BASE) algorithm that can learn the optimal neural network architectures for an entire task distribution simultaneously. The algorithm derived from a novel Bayesian view of architecture search utilizes the optimization embedding technique~\citep{DaiDaiHeLiuetal18} to automatically incorporated the task information into the parameterization of the posterior. We demonstrate the algorithm by training a meta-network simultaneous on a distribution of $2.634 \times 10^{23}$ tasks derived from \texttt{Imagenet} and achieve state-of-the-art results given our search time on both \texttt{CIFAR10}, \texttt{SVHN}, and \texttt{Imagenet} with quick adapted task-specific architectures. This work paves the way for future extensions with Meta Architecture Search such as direct fast-adaption to derive both optimal task-specific architectures and optimal weights and demonstrates the great efficiency gains possible by conducting architecture search over task distributions.
	
	\vspace{4mm}
	\subsubsection*{Acknowledgments}
We would like to thank the anonymous reviewers for their comments and suggestions. Part of this work was done while Bo Dai and Albert Shaw were at Georgia Tech.
Le Song was supported in part by NSF grants CDS\&E-1900017 D3SC, CCF-1836936 FMitF, IIS-1841351, SaTC-1704701, and CAREER IIS-1350983.
\clearpage
\newpage

	{
		\bibliographystyle{plain}
		\small
		\bibliography{./bibfile}
	}
	
	\clearpage
	\newpage
	
	\appendix
	\onecolumn

	\begin{appendix}
		
		\thispagestyle{plain}
		\begin{center}
			{\Large \bf Appendix}
		\end{center}
		\section{Architecture Space Details}\label{appendix:arch_space}
		For comparability in architectures, the particular search space used is very similar to that used in \citet{DARTS} and includes the same operation space: $3\times 3$, $5\times 5$, $7\times 7$ depth-wise separable convolutions, $3\times 3$ and $5\times5$
		dilated depth-wise separable convolutions, $3\times 3$ max pooling, $3\times 3$ average pooling, a $1\times 7$ followed by a $7\times 1$ convolution, skip connections, and no connection. In our search, each cell is made up of a total of six nodes with 2 input nodes. The input to each cell is the output from the previous 2 cells. The output for each cell is the concatenated output from all 4 non-input nodes in the cell. Following the same methods as \citet{DARTS, transferable}, non-dilated depth-wise separable convolutions were applied twice, all depth-wise separable convolutions did not have batch-norms between the grouped and 1x1 convolutions, convolutions had RELUs and batch-norms applied in ReLU-Conv-BN order, and all operations were padded as necessary to preserve spatial resolution as to only be reduced by the reducing layers whose first operations were applied with a stride of 2. 
		
		\subsection{\texttt{CIFAR10} and \texttt{Imagenet} Training Details}\label{appendix:retraining_details}
		
		\paragraph{\texttt{CIFAR10}}
		The architecture is transferred to a network with 20 cells following the motif shown in Appendix~\ref{appendix:single_task_arch}. The network was trained for 600 epochs with cutout augmentation. We used a batch size 96. We follow the same training strategy as \citet{DARTS} with cutout, and drop-path probability of 0.2, and auxiliary towers with weight 0.4.
		
		\paragraph{\texttt{SVHN}}
		The architecture is transferred to a network with 20 cells following the motif shown in Appendix~\ref{appendix:single_task_arch}. The network was trained for 160 epochs. We used a batch size 96, a drop-path probability of 0.2, and auxiliary towers with weight 0.4.
		The networks were trained for 160 epochs with cutout augmentation.
		
		\paragraph{\texttt{ImageNet}}
		The architecture is transferred to a network with 14 cells following the motif shown in Appendix~\ref{appendix:single_task_arch}. We train and evaluate in the mobile setting with input images of size 224x224. We train with a batch size of 256 for 375 epochs. We use the SGDR\citep{DBLP:journals/corr/LoshchilovH16a} learning rate schedule with $T_0=25$ and $T_mult = 2$. We optimize with the SGD with a initial lr of 0.1 decayed by a factor of 0.97 each epoch. We use a weight decay of $3e^{-5}$. For the remaing parameters we follow the same training strategy as \citep{transferable}.
		\newpage
		\subsection{Motifs for Single-Task Scalable Architectures}\label{appendix:single_task_arch}
		
		\begin{center}
			\begin{minipage}[b]{0.28\textwidth}
				\begin{center}
					\includegraphics[width=0.8\textwidth]{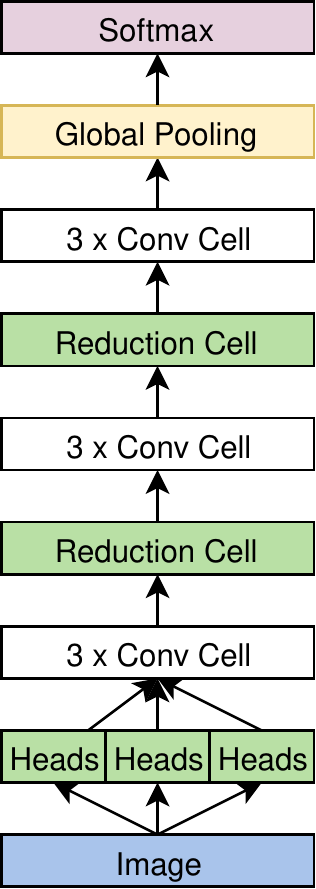}\\
					Motif for the Search \\
					Network
				\end{center}
			\end{minipage}
			\begin{minipage}[b]{0.28\textwidth}
				\begin{center}
					\includegraphics[width=0.8\textwidth]{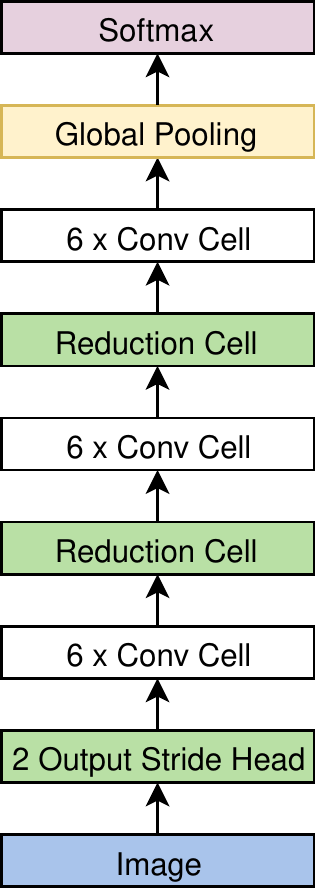}\\
					Motif for \texttt{CIFAR10} Full\\Network.
				\end{center}
			\end{minipage}
			\begin{minipage}[b]{0.28\textwidth}
				\begin{center}
					\includegraphics[width=0.8\textwidth]{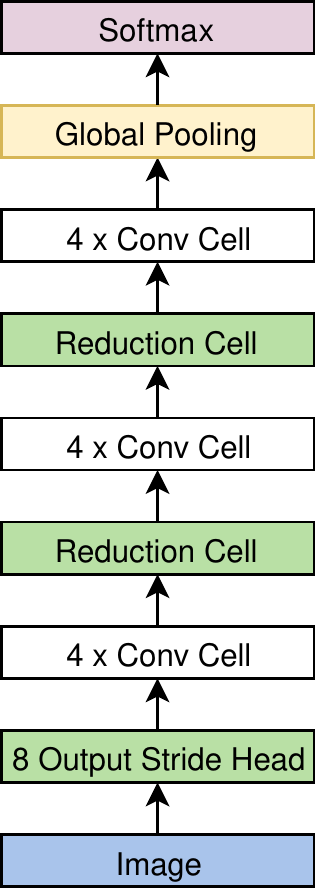}\\
					Motif for \texttt{ImageNet} Full\\Network.
				\end{center}
			\end{minipage}
		\end{center}
		
		These are the network motifs used in the experiments for search over single-task networks. Our search space has two unique cell architectures, "Normal Conv" and "Reduction" Cells.
		
		\subsection{Sample \texttt{ImageNet} Adapted Cell Designs}
		\vspace{-2mm}
		\begin{center}
			\begin{minipage}[b]{0.45\textwidth}
				\begin{center}
					\includegraphics[width=1.0\textwidth]{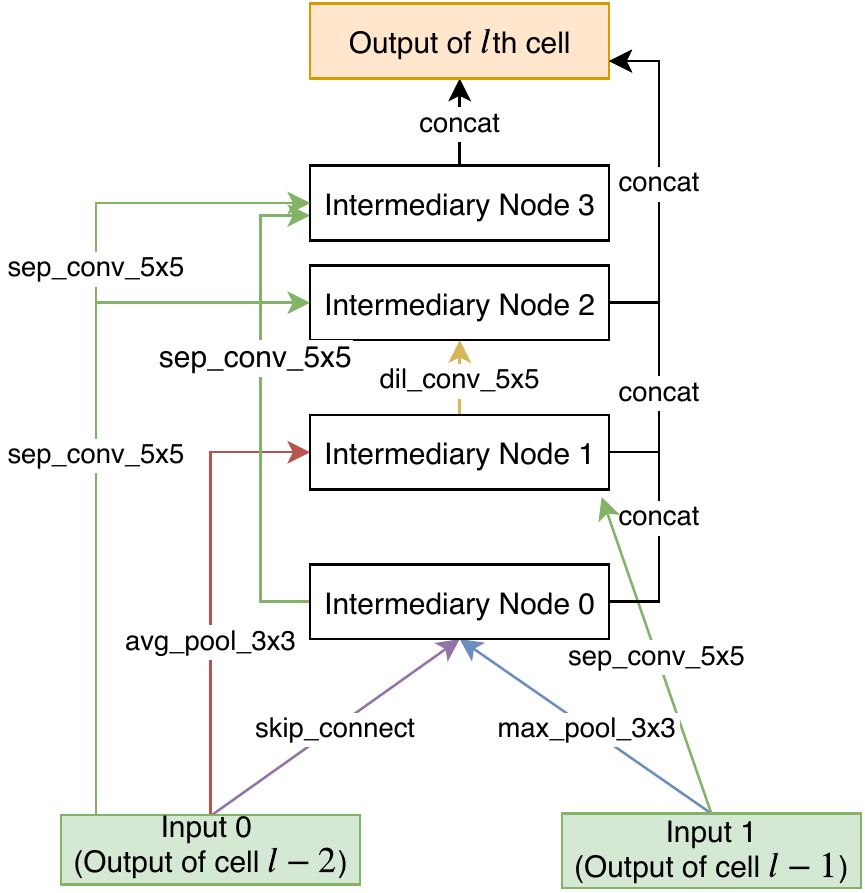}\\
					Cell Design for normal cell
				\end{center}
			\end{minipage}
			\hspace{1cm}
			\begin{minipage}[b]{0.45\textwidth}
				\begin{center}
					\includegraphics[width=1.0\textwidth]{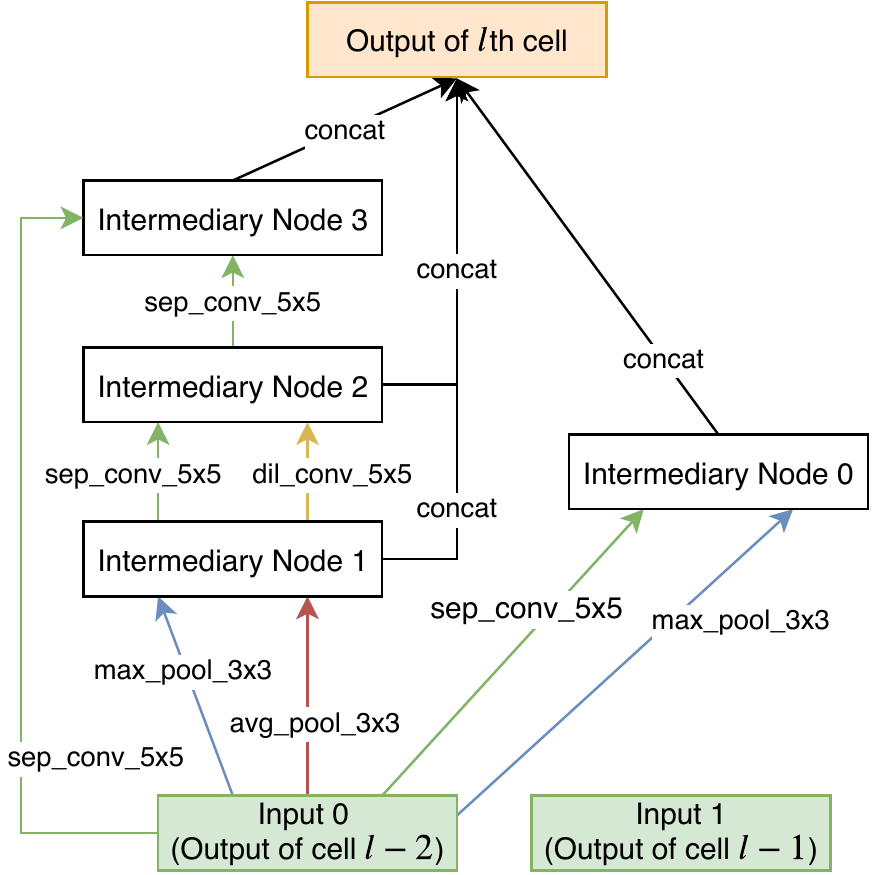}\\
					Cell Design for reduction cell
				\end{center}
			\end{minipage}
		\end{center}
		
		\section{Few Shot Learning}
		
		\subsection{Motifs for Scalable Architectures}\label{appendix:few_shot_arch}
		\begin{center}
			\begin{minipage}[b]{0.4\textwidth}
				\begin{center}
					\includegraphics[width=0.4\textwidth]{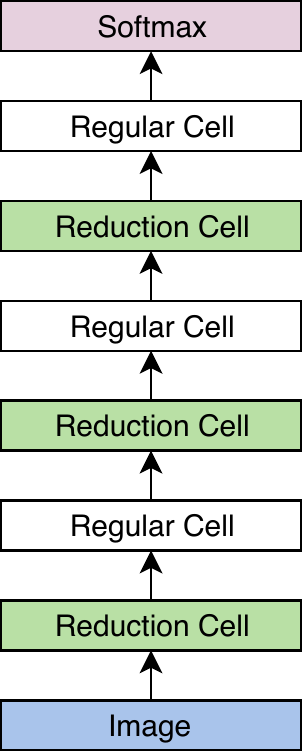}\\
					Motif for Search Network
				\end{center}
			\end{minipage}
			\begin{minipage}[b]{0.4\textwidth}
				\begin{center}
					\includegraphics[width=0.4\textwidth]{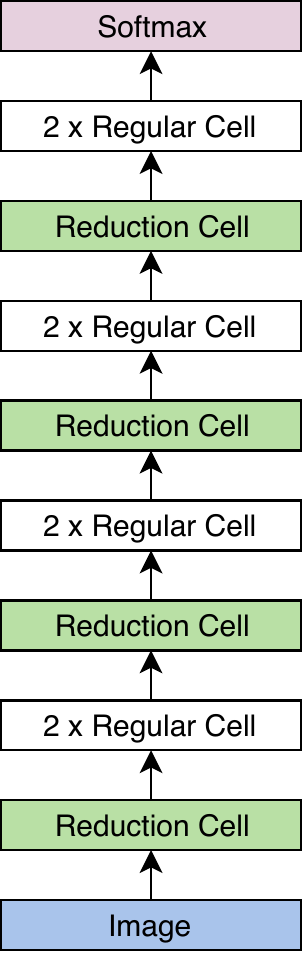}\\
					Motif for Full Network.
				\end{center}
			\end{minipage}
		\end{center}
		
		These are the network motifs used in the experiments for search over few-shot learning. Our search space has two unique cell architectures, "Normal" and "Reduction" Cells. The Meta Architecture Search was run with the "Search Network", and then for evaluation, the architectures were transferred to the full network.
		
		\subsection{High Level Diagrams of the Meta Architecture Search method.} \label{appendix:meta_arch_search_diagram}
		\begin{center}
			\begin{tabular}{cc}
				\includegraphics[width=0.35\linewidth]{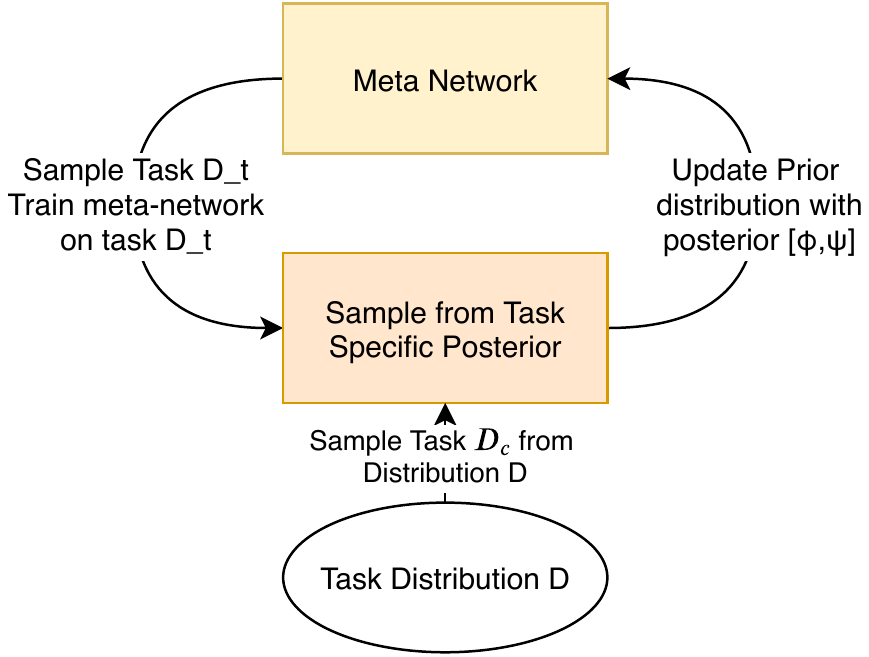} &
				\includegraphics[width=0.35\linewidth]{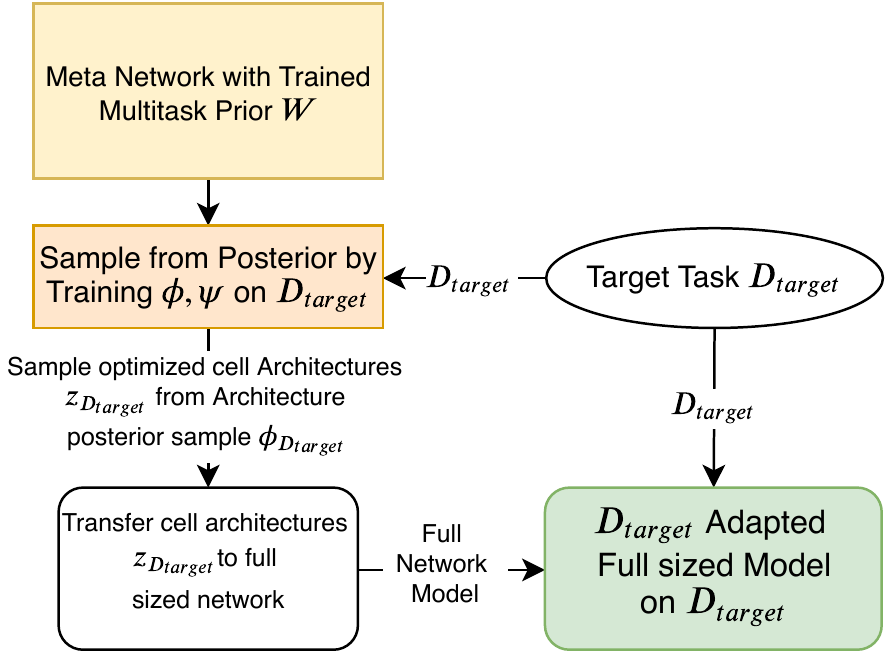}\hspace{20mm}\\
				(a) Meta Architecture Search & (b) One-Shot Architecture Adaptation\\
			\end{tabular}
		\end{center}
		
		\subsection{Diagram of Cell space concept}
		\label{appendix:cell_parameters}
		{
			\centering
			\includegraphics[width=0.8\textwidth]{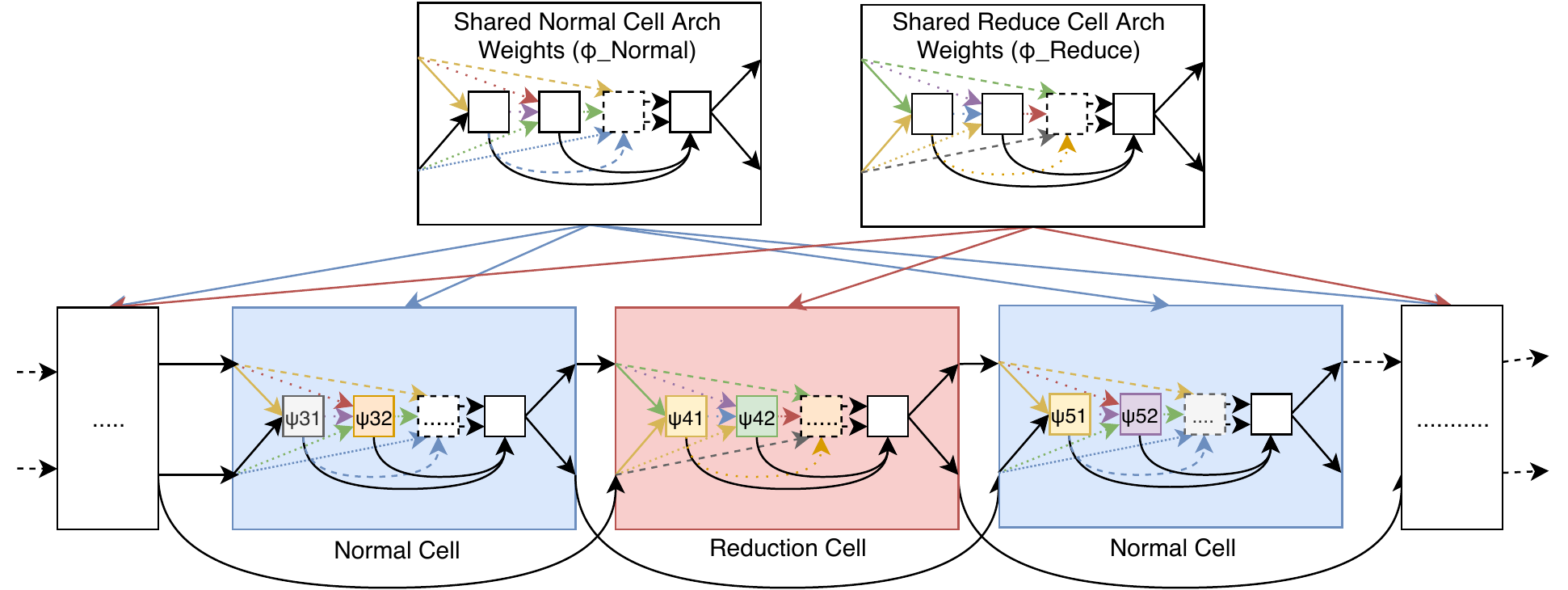}\\
		}
		The architecture parameters $\phi_{Normal}$ are shared between all architecture "normal cells" and describe the architecture distribution within in the cells.  $\phi_{Reduce}$ are shared between all reduce cells. All $\psi$ weight parameters are not unique to each layer.

		\subsection{Few-shot Training Details}\label{appendix:fewshot_training}
		In our experiments on the \texttt{Mini-Imagenet}  dataset, only the 64 training classes were used during training. The 12 validation classes were ignored, and evaluation was conducted on the 24 testing classes. 
		Search was run for $10000$ iterations. For each iteration, the meta-network was updated with the combined gradients from $T=2$ randomly sampled tasks. For each task $N=4$ steps of inner optimization were run. For the full training, all network architectures were trained with the same setting on the $5$-shot learning problem using the second-order MAML algorithm~\citep{MAML}. The full training was run for $30000$ iterations. Similarly, for each iteration, the network was again updated with the combined gradients from $2$ randomly sampled tasks, but each task was optimized with $5$ steps of inner optimization for second-order MAML. 
		
		\subsection{Sample Top Found Cell Architectures from few-shot BASE search}\label{appendix:cell_diagram}
		\begin{center}
			\begin{minipage}[b]{0.40\textwidth}
				\begin{center}
					\includegraphics[width=1.0\textwidth]{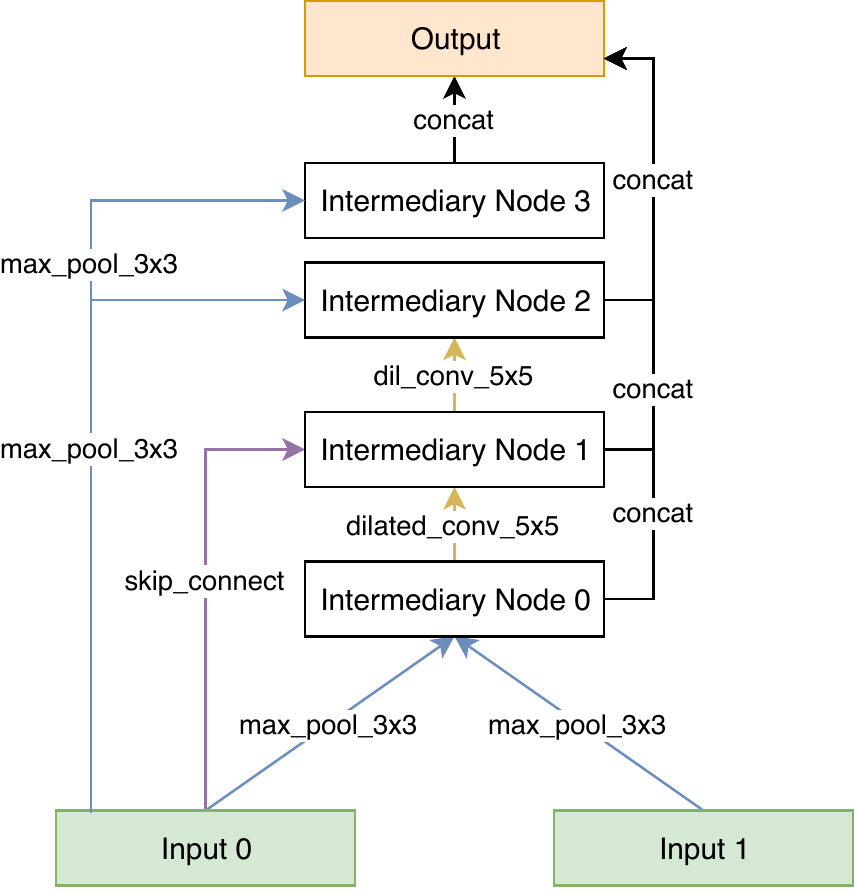}\\
					Cell Design for sample normal cell
				\end{center}
			\end{minipage}
			\hspace{1cm}
			\begin{minipage}[b]{0.4\textwidth}
				\begin{center}
					\includegraphics[width=1.0\textwidth]{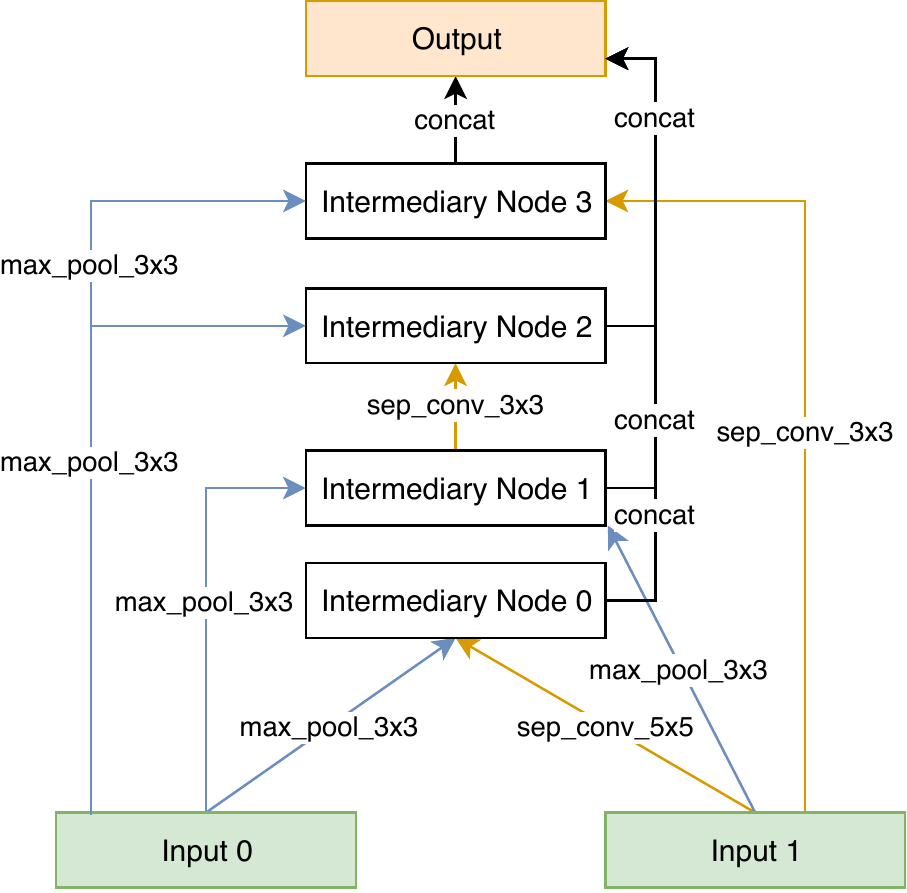}\\
					Cell Design for sample reduction cell
				\end{center}
			\end{minipage}
		\end{center}
		
	\end{appendix}
\end{document}